\documentclass[graybox]{svmult}

\usepackage{mathptmx}       
\usepackage{helvet}         
\usepackage{courier}        
\usepackage{type1cm}        
\usepackage{makeidx}         
\usepackage{graphicx}        
\usepackage{multicol}        
\usepackage[bottom]{footmisc}
\usepackage{xspace}

\usepackage[font={small},labelfont=bf,labelsep=space]{caption}
\usepackage{subcaption}
\usepackage{sidecap}
\usepackage{algorithmic}
\usepackage{algorithm}
\usepackage{amssymb}
\usepackage{amsmath}

\usepackage{array}
\newcolumntype{L}[1]{>{\raggedright\let\newline\\\arraybackslash\hspace{0pt}}m{#1}}
\newcolumntype{C}[1]{>{\centering\let\newline\\\arraybackslash\hspace{0pt}}m{#1}}
\newcolumntype{R}[1]{>{\raggedleft\let\newline\\\arraybackslash\hspace{0pt}}m{#1}}

\DeclareMathAlphabet{\mathcal}{OMS}{cmsy}{m}{n}

\newcommand{\pump}{PUMP\xspace}

\newcommand{\mpap}{MPAP\xspace} 
\newcommand{\pmp}{PMP\xspace} 
 
\newcommand{\phmp}{PHMP\xspace}

\newcommand{\xfree}{\mathcal X_{\text{free}}}
\newcommand{\xobs}{\mathcal X_{\text{obs}}}
\newcommand{\xgoal}{\mathcal X_{\text{goal}}}
\newcommand{\xinit}{x_{\mathrm{init}}}
\newcommand{\Psoln}{P_{\xgoal}}

\newcommand{\dr}{\lambda}

\newcommand{\Popen}{P_{\mathrm{open}}}

\newcommand{\B}{\mathcal{G}} 
\newcommand{\phtarget}{\beta}
\newcommand{\N}{\mathbb{N}}

\newcommand{\union}{\cup}
\newcommand{\intersection}{\cap}

\usepackage{eqparbox}

\newcommand{\LINEIF}[2]{%
    \STATE\algorithmicif\ {#1}\ \algorithmicthen\ {#2} \algorithmicend\ \algorithmicif }

\graphicspath{{./fig/}}
\usepackage{anyfontsize}

\renewcommand{\baselinestretch}{0.96}

\usepackage{titlesec}
\titlespacing*{\subsection}{0pt}
	{1\baselineskip plus 0.3\baselineskip minus 0.3\baselineskip}
	{1\baselineskip plus 0.3\baselineskip minus 0.3\baselineskip}
\titlespacing*{\section}{0pt}
	{1\baselineskip plus 0.3\baselineskip minus 0.3\baselineskip}
	{1\baselineskip plus 0.3\baselineskip minus 0.3\baselineskip}

\makeindex             

\begin{document}

\title*{Perception-Aware Motion Planning via Multiobjective Search on GPUs}
\author{Brian Ichter, Benoit Landry, Edward Schmerling, and Marco Pavone}
\institute{
Brian Ichter$^1$, Benoit Landry$^1$, Edward Schmerling$^2$, Marco Pavone$^1$ \at Dept. of Aeronautics and Astronautics$^1$ and Institute for Computational and Mathematical Engr.$^2$, Stanford University, Stanford, CA 94305\\
\email{\{ichter,blandry,schmrlng,pavone\}@stanford.edu} 
\\
\\ The authors gratefully acknowledge Anirudha Majumdar for insightful discussions and Patrick Kao for his help in the development of the quadrotor experimental platform. This work was supported in part by Qualcomm, the King Abdulaziz City for Science and Technology (KACST), and NASA under the Space Technology Research Grants Program (Grant
NNX12AQ43G). The GPUs used for this research were donated by the NVIDIA Corporation.
}
%
%
\maketitle

\vspace{-2.2cm}

\vspace{0.15cm}

\abstract{
In this paper we describe a framework towards computing well-localized, robust motion plans through the perception-aware motion planning problem, whereby we seek a low-cost motion plan subject to a separate constraint on perception localization quality.
To solve this problem we introduce the Multiobjective Perception-Aware Planning (\mpap) algorithm which explores the state space via a multiobjective search, considering both cost and a perception heuristic.
This framework can accommodate a large range of heuristics, allowing those that capture the history dependence of localization drift and represent complex modern perception methods.
We present two such heuristics, one derived from a simplified model of robot perception and a second learned from ground-truth sensor error, which we show to be capable of predicting the performance of a state-of-the-art perception system.
The solution trajectory from this heuristic-based search is then certified via Monte Carlo methods to be well-localized and robust. 
The additional computational burden of perception-aware planning is offset by GPU massive parallelization.
Through numerical experiments the algorithm is shown to find well-localized, robust solutions in about a second.
Finally, we demonstrate \mpap on a quadrotor flying perception-aware and perception-agnostic plans using Google Tango for localization, finding the quadrotor safely executes the perception-aware plan every time, while crashing in over 20\% of the perception-agnostic runs due to loss of localization.
}

\section{Introduction}\label{sec:intro}

The ability of an autonomous robotic system to localize itself in an environment is fundamental to robust operation in the field. While recent advances in perception have enabled this capability in a wide range of settings, the associated uncertainty can still vary significantly during operation.
Often this is due to variations in environmental knowledge, particularly in environments derived through simultaneous localization and mapping, or due to intrinsic environmental qualities, e.g., texture or lighting. 
Despite this, navigation in these environments is often posed as a motion planning problem agnostic to localization, operating under the assumption that the environment is known, yet the ability of a robot to localize itself against and track a trajectory is independent of its context.

In this paper, we approach the perception-aware motion planning problem, whereby we seek low-cost motion plans subject to a constraint on perception localization quality, in order to identify robust, well-localized plans. 
Unfortunately, the complexity of modern perception techniques and their history dependence make it difficult to quantify localization uncertainty, both computationally and theoretically, and thus a key contribution of this work is to perform perception-aware motion planning with a multiobjective search on both cost and a \emph{perception heuristic} to identify promising motion plans.
By considering trajectory cost and perception independently, both the problem formulation and algorithm can admit a wide range of perception heuristics applicable to modern closed-loop SLAM methods, including heuristics learned from empirical measurements of perception error.
We propose a sampling-based algorithm to approximate globally optimal plans in complex environments, the additional burden of considering perception and planning in tandem is offset through algorithmic design for massive parallelization on GPUs.
We term this algorithm the Multiobjective Perception-Aware Planning (\mpap) algorithm. Although \mpap leverages a heuristic to estimate localization uncertainty, we note that if an accurate generative model of perception error is available, the robustness of plans may be fully certified using Monte Carlo (MC) methods.

\subsection{Related Work}\label{sec:related}
While research has traditionally focused on improving methods for planning and perception without consideration of their later coupling, several methods to fuse the two have been put forward.
One possible approach is to formally combine them in a partially observable Markov Decision Process (POMDP) framework \cite{KaelblingLittmanEtAl1998}\cite{KurniawatiHsuEtAl2008}.
This approach, however, has been shown to quickly become intractable as the complexity or dimension of the problem increases.
Many works have sought to reduce this computational load by applying search techniques to graphical-representations of the belief space.
For example, the Belief Roadmap (BRM) was introduced in \cite{PrenticeRoy2009}, which applies ideas from probabilistic roadmap methods to planning in the belief space. 
The Rapidly-exploring Random Belief Trees algorithm \cite{BryRoy2011} similarly adapts the rapidly-exploring random trees algorithm to planning in the belief space.
Along with the use of sampling-based approaches, there exists a large body of work (including the above) that lessens the computational load of belief space planning through local approximations, receding horizon techniques, and simplifying assumptions like linearity, Gaussian belief spaces, stabilization phases, and maximum likelihood observations
\cite{PrenticeRoy2009, 
BryRoy2011, 
BergPatilEtAl2012, 
PatilKahnEtAl2014, 
IndelmanCarloneEtAl2015, 
BergAbbeelEtAl2011, 
Agha-mohammadiAgarwalEtAl2016, 
PlattTedrakeEtAl2010}. 
While these works approach the problem of perception-aware planning in a principled manner, the required assumptions and computational load of tracking belief states throughout the search can limit their applicability on high-dimensional problems or problems with very complex, difficult-to-quantify perception, as is the case with many modern perception methods.
Our approach simplifies the problem by avoiding the computation of the belief state during the graph search through instead only considering a perception heuristic and verifying the final result with Monte Carlo. Monte Carlo verification is done using the best available perception system model, and as an MC method it is asymptotically exact up to the model fidelity.

The field of active perception also considers the coupling of planning and perception, where the research problem is focused on computing robot motions that help minimize the uncertainty of localization and mapping \cite{AloimonosWeissEtAl1988,ThrunBurgardEtAl2005,CadenaCarloneEtAl2016}. In contrast to these works, the problem addressed in this paper adds an additional objective of ensuring high quality motion plans with respect to a dynamics-based cost objective. Compared to the active perception literature, we exchange modeling fidelity in a robot's perception systems for computational tractability in directly considering the tradeoff between perception quality and motion plan cost.

Some of the most similar works to our own have approached the problem as perception-aware motion planning.
In \cite{SadatChutskoffEtAl2014} a perception heuristic to capture the feature-richness of an environment model is introduced and used within a rapidly-exploring random tree (RRT) framework as a bound on individual edge, i.e., local connection, quality.
Another work, \cite{CostanteForsterEtAl2017}, proposes an approach to incorporate photometric information into the planning process, which is used as part of a weighted cost metric with path cost in an RRT search.
Lastly, \cite{CarloneLyons2014} considers an uncertainty-constrained planning problem within a receding horizon framework to compute feasible plans with mixed-integer linear programming.
In contrast to these works, we pose the problem as a bound on localization uncertainty (through a heuristic) of the \emph{entire} trajectory.
This shifts the approach more towards global motion plans and allows the use of more general heuristic formulations that can capture the history dependence of localization and trajectory deviation, though the proposed heuristics in these works too can be used in this framework.
As a brief example, consider a robot executing a trajectory in a learned environment that relocalizes (i.e., performs loop closure) given enough learned features in view.
A weighted cost and localization uncertainty metric is unable to capture future decreases in localization uncertainty and an edge-wise or limited time horizon approach may reject trajectories that would soon relocalize or conversely allow trajectories created from semi-localized edges that is never capable of full relocalization and continues to drift.

Finally, we note that the concept of multiobjective search to balance uncertainty with plan cost has been considered previously in motion planning, notably in \cite{IchterSchmerlingEtAl2017}.
We utilize the same multiobjective search accelerated through massively parallel GPU implementation put forth by the Parallel Uncertainty-aware Multiobjective Planning (\pump) algorithm \cite{IchterSchmerlingEtAl2017}.
However, while \pump considers only uncertainties internal to the system, this work considers external uncertainties through perception.
Additionally, in order to handle the added complexity of modern perception solutions, such as the Google Tango we experimentally demonstrate on, we instead employ a perception-heuristic constraint, and consider it in the objective search.

\subsection{Statement of Contributions}\label{sec:contrib}
 
In this work, we present the Multiobjective Perception-Aware Planning (\mpap) algorithm, which computes low-cost, well-localized, robust motion plans through perception-aware planning.
Built on a sampling-based graphical representation of the state space, the algorithm performs a multiobjective search to identify a set of Pareto optimal motion plans (considering cost and a perception heuristic).
By considering cost and perception separately, the perception heuristic can consider the history-dependence of the problem while being general enough to mirror modern perception techniques. 
We present one such heuristic, in this case based on a trained neural network, that maps sensor and estimation outputs of a robotic system to distributions of localization error rates.
The lowest-cost trajectory contained in the Pareto optimal set can then be certified to be well-localized and robust through asymptotically exact Monte Carlo methods.
The entire algorithm is designed specifically to be massively parallel and thus offset the burden of perception-awareness through the computational breadth of GPUs.

We demonstrate the efficacy of \mpap through both numerical and physical experiments in mapped environments. 
The numerical experiments demonstrate both that perception-aware plans can be generated on the order of seconds and that these plans are significantly more robust than their perception-agnostic counterparts. 
The physical experiments were performed by a quadrotor flying with a Google Tango smartphone performing localization in a previously mapped room. 
The resulting trials compare perception-agnostic and perception-aware trajectories, finding that the perception-agnostic flights crash several times due to loss of localization while the perception-aware flights complete all trajectories safely.

\subsection{Organization}\label{sec:org}
The remainder of this work is organized as follows.
Section~\ref{sec:probstate} details the problem formulations considered herein.
Section~\ref{sec:alg} formally describes the \mpap algorithm.
Section~\ref{sec:numexp} demonstrates the algorithm's efficacy and timing over a series of simulations.
Section~\ref{sec:exp} demonstrates the algorithm and a learned heuristic through physical experiments, further motivating the need for perception-aware planning.
Lastly, Section~\ref{sec:conc} summarizes our findings and offers directions for future work.

\section{Problem Statement}\label{sec:probstate}

This work seeks to compute a trajectory from a start state to a goal region that is low-cost, dynamically-feasible, collision-free, and well-localized throughout its execution. 
We consider a robot with state space $\mathcal X \subset \mathbb{R}^d$ and dynamics $\dot{x} = f(x,u)$, where $x \in \mathcal{X}$ is the state and $u \in \mathbb{R}^\ell$ is the control input.
Let $\xobs$ be the obstacle space and $\xfree := \mathcal{X} \setminus \xobs$ be the free space, which the robot must navigate through to get from an initial state $\xinit \in \xfree$ to a goal region $\xgoal \subset \xfree$.
The robot tracks a planned trajectory $x^\text{nom}:[0,T] \to \mathbb{R}^d$ using a perception-based state estimate $\hat{x} \in \mathbb{R}^d$ and closed-loop controller.
In the process of tracking this trajectory, the robot is subject to state estimate noise due to imperfect measurements, resulting in a localization error $||\hat{x}(t) - x(t)||$ at time $t$.
Given this, we approach the perception-aware motion planning problem by seeking a minimum cost trajectory (with cost measure $c$) to be localized within a bound $\delta \hat{x}$ with high probability (1-$\alpha$).

\vbox{\noindent\textbf{Perception-aware Motion Planning (\pmp) problem:}
\vspace{-0.3cm}
\begin{equation}
\label{eq:pmp}
\begin{split}
\min_{u^\text{nom}(\cdot)} & \quad c(x^\text{nom}(\cdot)) \\
\text{s.t.} 
& \quad \dot{x}^\text{nom} = f(x^\text{nom},u^\text{nom}) \\
& \quad x^\text{nom} \intersection{} \xobs = \emptyset \\
& \quad x^\text{nom}(0) = \xinit, \quad x^\text{nom}(T) \in \xgoal \\
& \quad \mathbb{P}(\{||\hat{x}(t) - x(t)|| \mid t \in [0,T]\} \geq \delta \hat{x}) \le \alpha.\\
\end{split}
\end{equation}}
\vspace{-0.4cm}

Unfortunately, perceptual localization error can be difficult to quantify, particularly for entire trajectories.
Instead, we propose the use of a perception heuristic conditioned on the nominal trajectory and the environment as a proxy to the true localization error. 
We define the heuristic as $h(x^\text{nom}([0,t]),\mathcal{E})$, where $x^\text{nom}([0,t])$ denotes the planned trajectory over the time range $[0,t]$ and $\mathcal E$ the environmental information used for perception (e.g., visual features), and bound its value by $\beta$. \newline
\noindent\textbf{Perception-Heuristic-aware Motion Planning (\phmp) problem:}
\vspace{-0.3cm}
\begin{equation}
\label{eq:phmp}
\begin{split}
\min_{u^\text{nom}(\cdot)} & \quad c(x^\text{nom}(\cdot)) \\
\text{s.t.} 
& \quad \dot{x}^\text{nom} = f(x^\text{nom},u^\text{nom}) \\
& \quad x^\text{nom} \intersection{} \xobs = \emptyset \\
& \quad x^\text{nom}(0) = \xinit, \quad x^\text{nom}(T) \in \xgoal \\
& \quad h(x^\text{nom}([0,t]),\mathcal E) \le \phtarget \quad \forall t \in [0,T].\\
\end{split}
\end{equation}
\vspace{-0.4cm}

By considering \phmp as a cost optimization with a separate perception constraint, this formulation allows the heuristic to capture the history dependent nature of trajectory and localization drift and can be tailored to the available sensors and algorithmic approaches.
This heuristic need not be linear or monotonic, and in general should be tuned to the system's perception method and empirical results.
In Section~\ref{sec:learnedheur} we provide an additive, learned heuristic based on state and visual features to approximate the localization error rate, but many other possible heuristics exist.
Additional examples include estimated uncertainty covariance, observability, map entropy, or likelihood of loop closure in an area, and in general may incorporate complex evolution over state history as needed.

Ultimately, the goal of the perception-aware motion planning problem is to find a robust trajectory -- a trajectory that can be tracked closely with high probability.
Formally, we state the \emph{robust motion planning problem} as $\mathbb{P}(\{||x^\text{nom}(t) - x(t)|| \mid t \in [0,T]\} \geq \delta x^\text{nom}) \le \alpha$, where $x$ is the executed trajectory, $\delta x^\text{nom}$ is the deviation bound, and $\alpha$ is the probability bound.
While this deviation may result from many sources, including dynamics modeling errors, environmental disturbances, and process noise, we focus on localization error due to perception as one major source of failure in modern robotic systems.
In settings where localization error is the primary source of deviation, the solution to the \pmp problem may serve as a proxy to the robust motion planning problem.

\section{The Multiobjective Perception-Aware Planning Algorithm}\label{sec:alg}

In this section we introduce the Multiobjective Perception-Aware Planning (\mpap) algorithm.
The algorithm assumes an a priori mapped environment, including both obstacles and some sense of perceptual capabilities (e.g., locations of features), though we do present a methodology for use within a SLAM framework at the end of this section.
\mpap computes well-localized, robust motion plans initially through the perception-heuristic-aware motion planning problem via a multiobjective search of the state space to compute the Pareto optimal front of cost and a perception heuristic.
This search removes any Pareto dominated plans, as well as any plans in violation of the perception-heuristic constraint.
The lowest-cost feasible plan resulting from this search is then used to inform robustness through a Monte Carlo (MC) verification of the full system.
The resulting simulations can be used to iteratively update the perception-heuristic bound (either loosening or tightening) until a robust solution is found.
The additional computational load of a multiobjective search and a MC verification are offset through algorithmic design for massive parallelization to leverage the computational power of GPUs.
This approach is outlined in Alg.~\ref{alg:MPAPoutline} and thoroughly detailed below.
We note this algorithm uses the same multiobjective search approach as well as massive parallelization strategy introduced in \cite{IchterSchmerlingEtAl2017}.
In the interest of completeness, we detail the algorithm in its entirety here, while also providing new discussion relevant to perception-aware motion planning.

\begin{algorithm}
\caption{\mpap: Outline}
\label{alg:MPAPoutline}
\algsetup{linenodelimiter=}
\begin{algorithmic}[1]
\STATE Massively parallel sampling-based graph building (Alg.~\ref{alg:graphBuild})
\STATE Massively parallel perception heuristic computation
\STATE Multiobjective search to find optimal plan subject to a perception-heuristic constraint (Alg.~\ref{alg:explore})
\STATE Monte Carlo verification of localization and robustness, and refinement of bound for Alg.~\ref{alg:explore}
\end{algorithmic}
\end{algorithm}

We now present formal notation and algorithmic primitives to aid in the elucidation of \mpap.
We define sampled nodes (samples) in $\mathbb{R}^d$ where $d$ is the dimension of the state space.
We represent plans as structs $(\mathrm{head},\mathrm{path},\mathrm{cost},\mathrm{h})$, where each respective field represents the terminal trajectory node from which we may extend other plans, the list of previous nodes, the cost, and the perception heuristic.
In the following definitions let $u,v \in \xfree$ be samples and let $V \subset \xfree$ be a set of samples.
Define the following standard algorithmic primitives: $\texttt{SampleFree}(n)$ returns a set of $n \in \N$ points sampled from $\xfree$, $\texttt{Cost}(u,v)$ returns the cost of the optimal trajectory from $u$ to $v$,
$\texttt{Near}(V,u,r)$ returns the set of samples $\{v \in V : \texttt{Cost}(u,v) < r\}$, and
$\texttt{Collision}(u,v)$ returns a boolean value indicating whether the optimal trajectory from $u$ to $v$ intersects $\xobs$.

Further define the following functions specific to the \mpap algorithm.
Given a set of motion plans $P$ and $\Popen \subset P$, $\texttt{RemoveDominated}(P,\Popen)$ denotes a function that removes all $p \in \Popen$ from $\Popen$ and $P$ that are dominated by a motion plan in $\{p_\text{dom} \in P : p_\text{dom}.\text{head} = p.\text{head}\}$; we say that $p_\text{dom}$ dominates $p$ if $(p.\text{cost} > p_\text{dom}.\text{cost}) \wedge (p.\text{h} \geq p_\text{dom}.\text{h})$.
Define  $\texttt{PH}(v,p)$ as a function that returns the value of the perception heuristic for the motion plan $p$ concatenated with the optimal connection $p.\text{head}$ to $v$.

With these algorithmic primitives and notation in hand, we now present the full \mpap algorithm.
Note that for brevity, as well as clarity, any necessary tuning parameters or variables not explicitly passed to functions are assumed to be global.
Given a mapped environment, the algorithm begins with a graph building stage (Alg.~\ref{alg:graphBuild}) to construct a representation of available motions through the state space, similar to probabilistic roadmap methods \cite{KavrakiSvestkaEtAl1996}.
The graph is initialized with $\xinit$ and $n$ samples from $\xfree$, via $\texttt{SampleFree}(n)$, to form the set $V$.
This set of samples must include at least one from $\xgoal$.
An $r_n$-disc graph is then formed by computing the neighbors of each sample within an $r_n$ cost radius and checking their connection is collision free.
The value of $r_n$ can be considered a tuning parameter in line with standard practices in sampling-based planning.
For kinodynamic systems, these edges should consider the respective kinematics and dynamics.
This first stage represents an embarrassingly parallel process following the discussion in \cite{AmatoDale1999}.

\begin{algorithm}[h!]
\caption{\texttt{BuildGraph}}
\label{alg:graphBuild}
\algsetup{linenodelimiter=}
\begin{algorithmic}[1]
\STATE $V \leftarrow \{\xinit\} \union \texttt{SampleFree}(n)$
\FORALL[massively parallel graph building]{$v \in V$}
\STATE $N(v) \leftarrow \texttt{Near}(V \setminus \{v\}, v, r_n)$
\FORALL{$u \in N(v)$}
\LINEIF{$\texttt{Collision}(v,u)$}{$N(v) \leftarrow N(v) \setminus \{u\}$}
\ENDFOR
\ENDFOR
\RETURN ($V$, $N$)
\end{algorithmic}
\end{algorithm}

With the graph complete, the second phase performs any computation of the perception heuristic that does not require full trajectories.
Key to the computational tractability of this algorithmic approach is the choice of a perception heuristic amenable to significant, massively parallel computation before the full trajectory is determined in the graph search.
This may be a heuristic that is derived through the combination of nodes visited or edges checked, allowing the value of each individual section to be computed independently on the GPU.
We present two such heuristics in Sections~\ref{sec:numexp} and \ref{sec:exp}, but note that the heuristic can be quite general and should be set to the given perception method.

Following the graph construction and any massively parallel computation for the perception heuristic, a massively parallel multiobjective search is performed over the state space, considering both the trajectory cost and perception heuristic, as detailed in Alg.~\ref{alg:explore}.
The exploration proceeds by iteratively expanding groups, $\B$, in parallel to their nearest neighbors.
These groups are formed from the set of open motion plans, $\Popen$, below an increasing cost threshold defined for each iteration $i$ as $i \dr r_n$, where $\dr \in (0, 1]$ denotes the group cost factor.
The value of $\dr$ is a tuning parameter that represents a tradeoff between parallelism and the potential for discarding promising plans or early termination.
For higher values of $\dr$ the number of samples in each group increases, and thus the computation is made more parallel, but this comes at the cost of spreading the set $\Popen$ (as plans are left behind the cost threshold), which potentially allows the search to terminate before the optimal motion plan has reached $\xgoal$.
At the end of each iteration any dominated plans are removed, as are any plans found in violation of the perception-heuristic constraint ($p.\text{h} > \phtarget$).
The exploration terminates when a feasible plan reaches $\xgoal$ or $\Popen = \emptyset$.
If multiple plans are found in the Pareto front that satisfy the constraint bound, the minimum cost plan is selected.
We note that in addition to the parallelism introduced over individual plans, the use of a cost bound, $i \dr r_n$, allows for plans to be stored in cost thresholded buckets with $\mathcal{O}(1)$ insertion and removal.

\begin{algorithm}[h]
\caption{\texttt{Explore}}
\label{alg:explore}
\algsetup{linenodelimiter=}
\begin{algorithmic}[1]
\STATE $\Popen \leftarrow \{(\xinit, \emptyset, 0, 0)\}$    \hspace{3.9cm} // plans ready to be expanded
\STATE $P(\xinit) \leftarrow \Popen$                        \hspace{4.8cm} // plans with head at $\xinit$
\STATE $\B \leftarrow \Popen$                               \hspace{5.45cm} // plans considered for expansion
\STATE $i = 0$
\WHILE{$\Popen \neq \emptyset \land \{g \in \B : (g.\mathrm{head} \in \xgoal) \wedge
    (g.\mathrm{h} \leq \phtarget)\} = \emptyset$}\label{line:exploreWhile}
\FORALL[in parallel $\qquad\qquad\qquad\:\:\quad\qquad$]{$p \in \B$}                      
\FORALL[in parallel $\qquad\qquad\qquad\:\:\quad\qquad$]{$x \in N(p.\mathrm{head})$}         
\STATE $q \leftarrow (x, p.\mathrm{path} + \{p.\mathrm{head}\}, p.\mathrm{cost} + \texttt{Cost}(p.\mathrm{head},x),\texttt{PH}(x,p))$\label{line:newPath}
\IF[perception-heuristic cutoff $\qquad\:\:\:\:$]{$q.\mathrm{h} \leq \phtarget$}
\STATE $P(x) \leftarrow P(x) \union \{q\}$
\STATE $\Popen \leftarrow \Popen \union \{q\}$
\ENDIF
\ENDFOR
\ENDFOR
\STATE $(P,\Popen) \leftarrow \texttt{RemoveDominated}(P,\Popen)$
\STATE $\Popen \leftarrow \Popen \setminus \B$
\STATE $i \leftarrow i + 1$
\STATE $\B \leftarrow \{p \in \Popen : p.\mathrm{cost} \leq i \dr r_n\}$
\ENDWHILE
\STATE $P_\mathrm{candidates} \leftarrow \{p \in P(v) : v \in \xgoal\}$
\RETURN $\Psoln \leftarrow \mathrm{min}_{p \in P_{\mathrm{candidates}}}\{p.\mathrm{cost}\}$
\end{algorithmic}
\end{algorithm}

\begin{figure}[b]
\sidecaption
    \centering
    \includegraphics[width=0.5\textwidth]{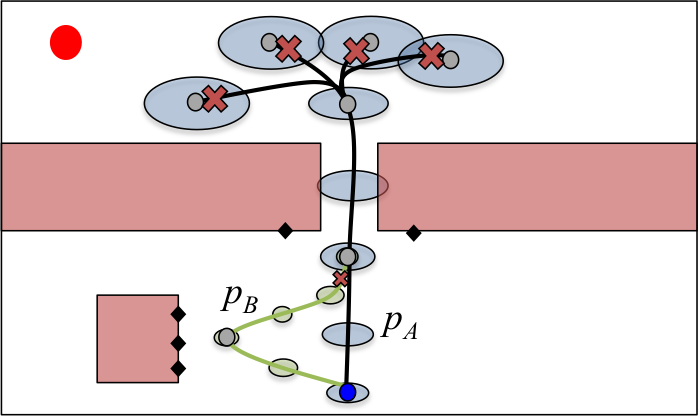}
    \caption{An uncertainty-constrained planning problem in which a multiobjective formulation should be used over a weighted cost metric. 
Features to localize against are shown as black diamonds. 
While the partial plan $p_A$ may outperform $p_B$ in a combined cost metric, it eventually exceeds the constraint. A multiobjective search will consider continuations of both plans, while a combined formulation may extend only one.
}
\label{fig:multiobjective}
\end{figure}

As discussed, we employ a multiobjective search to solve the \phmp problem, but it may be natural to ask why incur the computational burden of a multiobjective search instead of a combined cost function.
Since we are approaching a problem that considers an objective and constraint, we should consider both independently as we do not know how they will progress in the future.
For example, consider the two potential plans expanding through a narrow passageway in Fig.~\ref{fig:multiobjective}.
Plan $p_A$ has a low cost, medium constraint value (but has not yet violated the constraint), and low combined cost function, while plan $p_B$ has a high cost, low constraint value, and high combined cost function.
With our combined cost function, $p_A$ clearly dominates and is expanded through the narrow passageway, spawning several other plans as concatenations with $p_A$.
If all of these new plans incur high additional penalties on the constraint value, none will be valid.
Unfortunately, because the narrow passageway is already filled by $p_A$, the potentially feasible plans that begin with $p_B$ are never realized.
Though a different weighted combination could be used, there is no correct choice for arbitrary problems.
The multiobjective search has the further benefit of admitting a wide range of perception heuristic functions. In this work we consider additive heuristics intended to emulate the variance in a localization estimate (indeed, heuristics of this form could be addressed by state augmentation in a non-sampling-based context) but, e.g., more complicated particle-based heuristics
have also been applied within a similar multiobjective search framework \cite{IchterSchmerlingEtAl2017}.

The final phase of \mpap computes an asymptotically exact probability of motion plan $p$ satisfying a localization error bound through MC sampling \cite{JansonSchmerlingEtAl2015b}.
Given the output of this search, the perception-heuristic bound may be updated (increased to allow more aggressive trajectories or decreased for more conservatism) and the exploration rerun.
Once a final feasible plan has been determined, this step gives a certificate of constraint satisfaction.
This final phase too can be augmented by a smoothing algorithm to adjust the pose along the trajectory, subject to feasibility, to maximize perception (one potential approach is outlined in \cite{DavisonMurray2002}).

Briefly, we note that this final phase can be considered optional, dependent on the ability to accurately model the system's trajectories and perception as well as the quality of the perception heuristic used.
Despite the flexibility of the MC approach, in some cases due to complex dynamics, modeling the system's trajectories may be computationally intractable or the onboard perception may not be well known, rendering a MC step inaccurate.
We demonstrate this case in Section~\ref{sec:exp} of this work, where a Google Tango smartphone is used for localization, which, to the authors, may be considered a black box localization framework.
Without the MC verification, significant gains are still observed by simply considering the perception heuristic.

\emph{Online implementation:} Finally, we discuss here implementing \mpap for real-time planning in conjunction with SLAM.
The computational load of the \mpap algorithm is primarily in the graph building and computation of the perception heuristic (for a wide range of choices), whereas the multiobjective search is relatively lightweight.
Thankfully, both of these intensive algorithms can easily be iteratively updated as new information becomes available.
For the graph building, if new obstacles are discovered or regions are updated, the number of new edge collision checks can be limited.
For the perception heuristic, if updates are local (such as a new region or feature's discovery), again only small changes need occur.
The exploration loop, being separate algorithmically from the other two, can then be rerun as needed.

\section{Numerical Experiments}\label{sec:numexp}

\subsection{Simulation Setup}\label{sec:simsetup}

In this section we demonstrate the performance of the \mpap algorithm through numerical simulations with a quadrotor-inspired dynamical system operating in a known 3D workspace with visual-inertial odometry (VIO) localization.
We model our system as a 6D double integrator ($\ddot{x} = u$) with an additional single integrator yaw state.
The system's dynamics are propagated by discretized approximate dynamics and the nominal trajectory is tracked by an LQR trajectory tracking controller.
For simplicity, the yaw state is considered to be tracked exactly.

The system is localized in MC through a learned-feature-based VIO method.
The inertial estimate is obtained through a simulated accelerator with noise.
The visual localization estimate is based on matching viewed features with known 3D locations (up to a level of noise), as to emulate a learned environment.
At each timestep the system identifies the features in the field of view (unobstructed by an obstacle and within a field of view angle from the system's yaw direction) and is given a relative position with noise, as may be estimated from motion and successive images (for simplicity is assumed that features are identified perfectly if in the field of view).
The position is then found from these features using a 3D-to-3D correspondence method described in \cite{ScaramuzzaFraundorfer2011}.
Finally the estimates are fused via a Kalman filter.

For the numerical simulations in this section we introduce a simple feature-based perception heuristic, matching our system's localization method (in Section~\ref{sec:exp} we present a more generally applicable learned heuristic).
At each timestep $dt$ along a given trajectory, an additive penalty of $dt$ is incurred to represent increasing drift.
This penalty is offset when features are in view (i.e., in the field of view and unobstructed) by decrementing the heuristic by $dt/n_f$, where $n_f$ represents the number of features required to offset any drift.
If the number of features in view exceeds $n_f$, the heuristic will decrease in time, representing possible relocalization (loop closure).
We set $n_f = 12$ for the remainder of this work.
If the heuristic becomes negative, it is set to zero to represent the concept that a robot can at most be fully localized.
This heuristic is able to capture both the character of our problem and the history-dependence of localization drift,
and demonstrates the capability of the \mpap framework.

\subsection{Simulation Results}\label{sec:simresults}

The simulations were implemented in CUDA C and run on an NVIDIA GeForce GTX 1080 Ti GPU on a Unix system with a 3.0 GHz CPU.
Our implementation samples the state space using the deterministic, low-dispersion Halton sequence.
As this set of samples is deterministic, no variance results are presented.
Leveraging this determinism, we perform an offline precomputation phase to compute both nearest neighbors and edge discretizations.
We also note that the obstacles used in our simulations represent axis-aligned bounding boxes, as often used in a broadphase collision checking phase \cite{LaValle2006}.
This methodology can provide increasingly accurate representations of obstacles as needed.
Finally, we set $\dr = 0.5$, which we have found represents a good tradeoff between parallelism, ease of implementation, and potential early termination.

\begin{figure}[t]
    \centering
    \begin{subfigure}[b]{0.31\textwidth}
        \includegraphics[width=\textwidth]{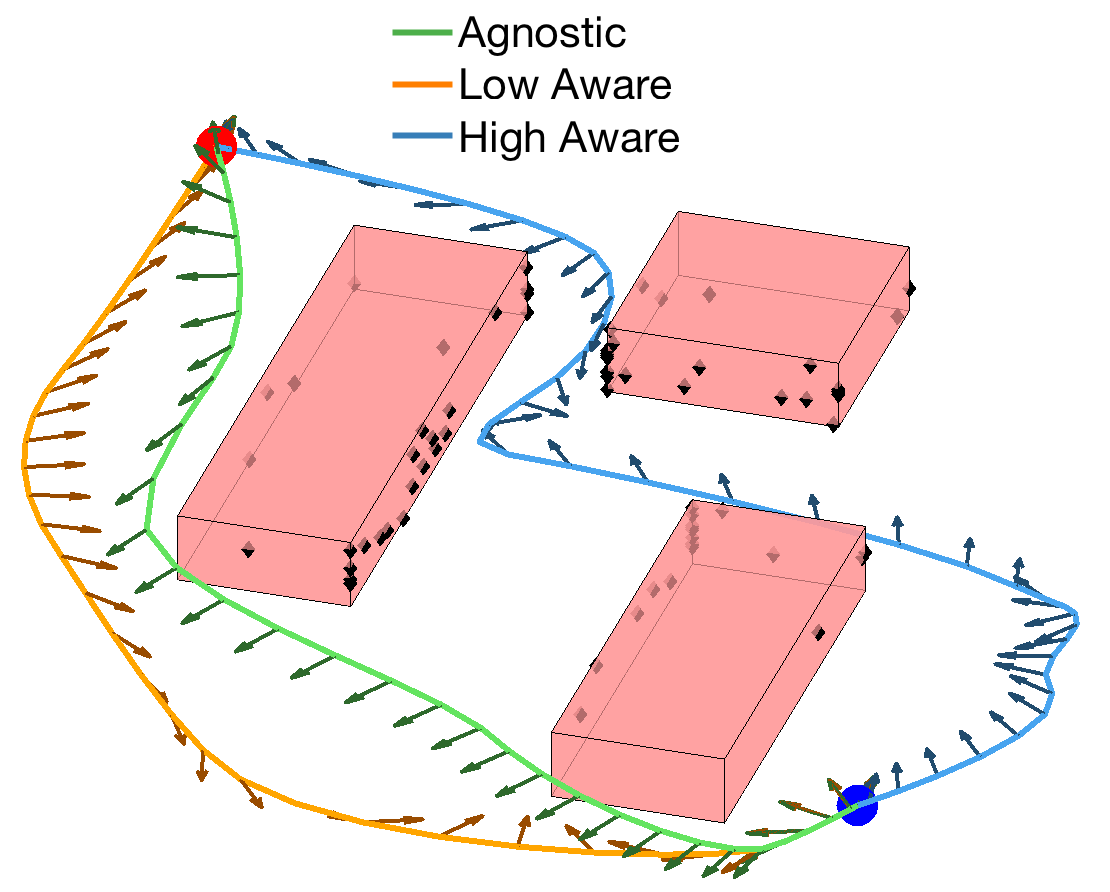}
        \caption{}
        \label{fig:inputMap}
    \end{subfigure}
    \begin{subfigure}[b]{0.315\textwidth}
        \includegraphics[width=\textwidth]{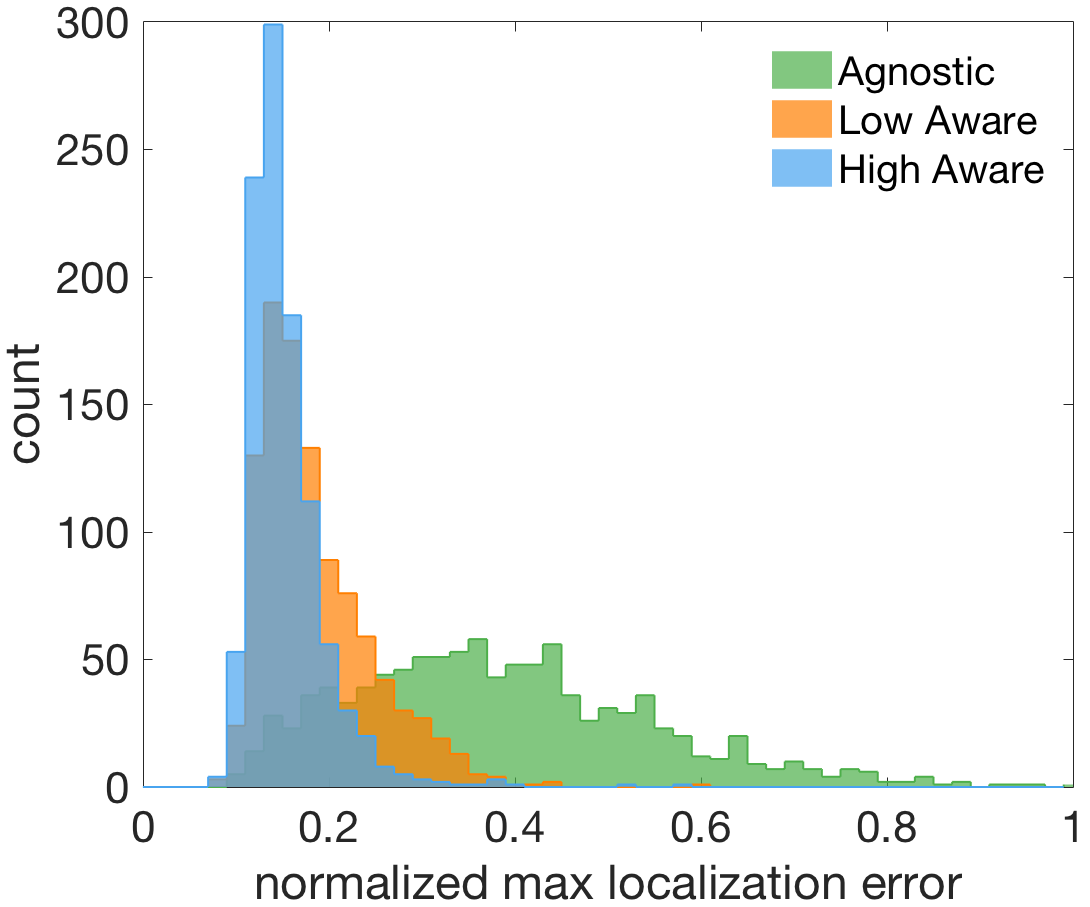}
        \caption{}
        \label{fig:inputHistEst}
    \end{subfigure}
    \begin{subfigure}[b]{0.32\textwidth}
        \includegraphics[width=\textwidth]{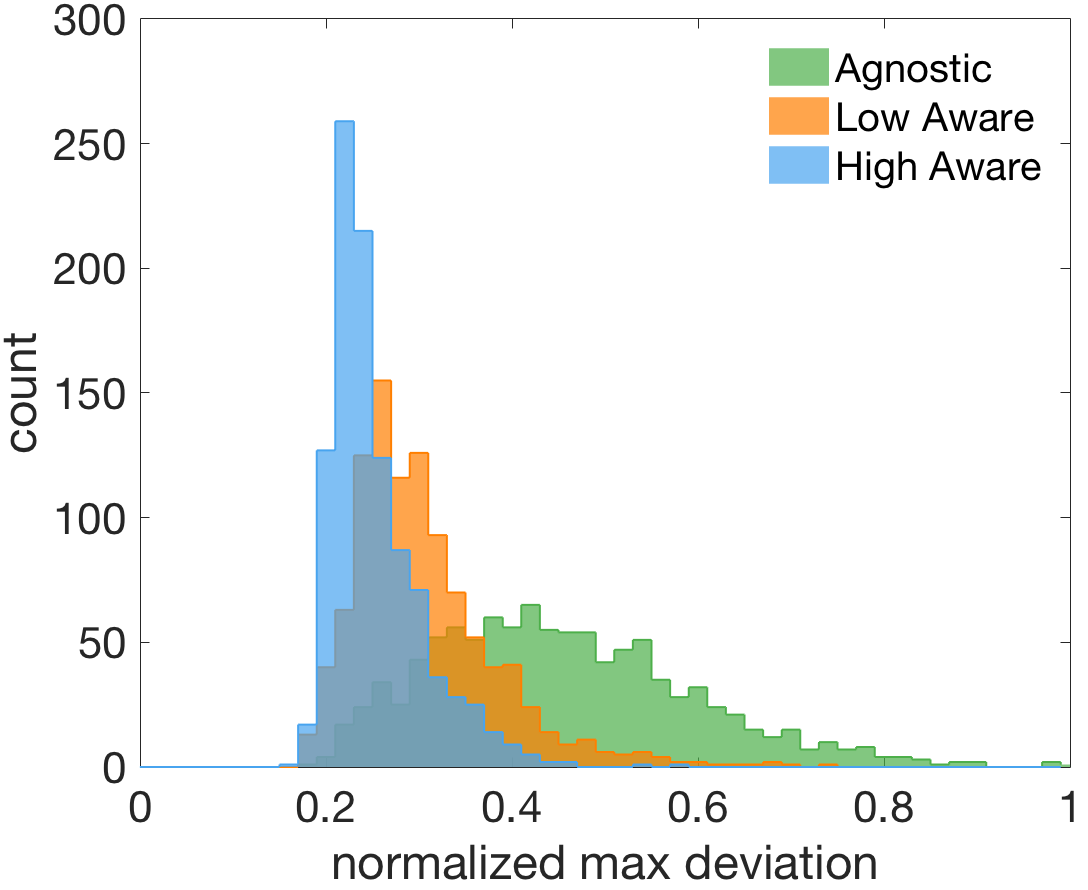}
        \caption{}
        \label{fig:inputHist}
    \end{subfigure}
\caption{
(\ref{fig:inputMap}) \mpap motion plans with varying levels of perception heuristic bounds. Mapped features are shown in black, but are only visible for localization if within the field of view and unobstructed. The low aware trajectory improves over the agnostic trajectory primarily with changes in yaw, whereas for high awareness the solution homotopy class must change. 
(\ref{fig:inputHistEst}, \ref{fig:inputHist}) The histograms of maximum localization error and trajectory deviation from MC simulations demonstrates perception-awareness results in both lower mean deviations and less variance.}
\label{fig:input}
\end{figure}

The first simulation environment, shown in Fig.~\ref{fig:input}, is rather simple, but demonstrates the benefits of perception-aware planning.
The obstacle set was generated to allow for low-cost trajectories around the outside of the workspace, but with a feature distribution favoring the inside of the workspace.
Three plans were generated for varying levels of perception heuristic (from highly perception-aware to perception-agnostic).
As can be seen, the perception-agnostic and low perception-aware plans utilize the same solution homotopy, varying primarily in yaw direction.
The highly perception-aware plan however changes homotopy class.
Figs.~\ref{fig:inputHistEst}-\ref{fig:inputHist} and Table~\ref{table:costDev} show the MC simulation results, which demonstrate the robustness benefits both in terms of mean, but also variance.
The distributions vary most significantly in the length of the right tails, which is the regime in which robustness is most impacted.
The errors for the top 1\% of plans are shown in Table~\ref{table:costDev}.

\begin{figure}[h!]
    \centering
    \begin{subfigure}[b]{0.39\textwidth}
        \includegraphics[width=\textwidth]{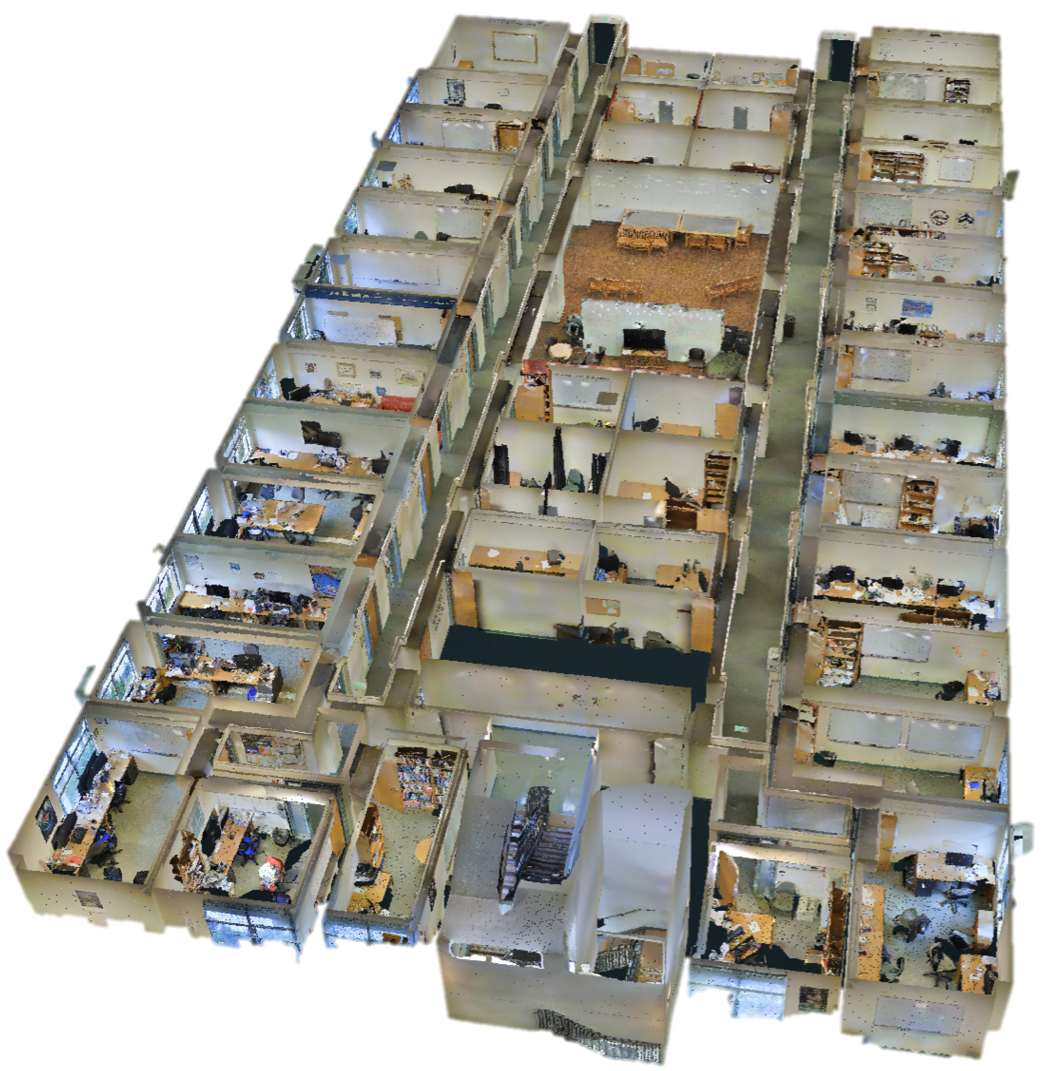}
        \caption{}
        \label{fig:gatesReal}
    \end{subfigure}
    \begin{subfigure}[b]{0.28\textwidth}
        \includegraphics[width=\textwidth]{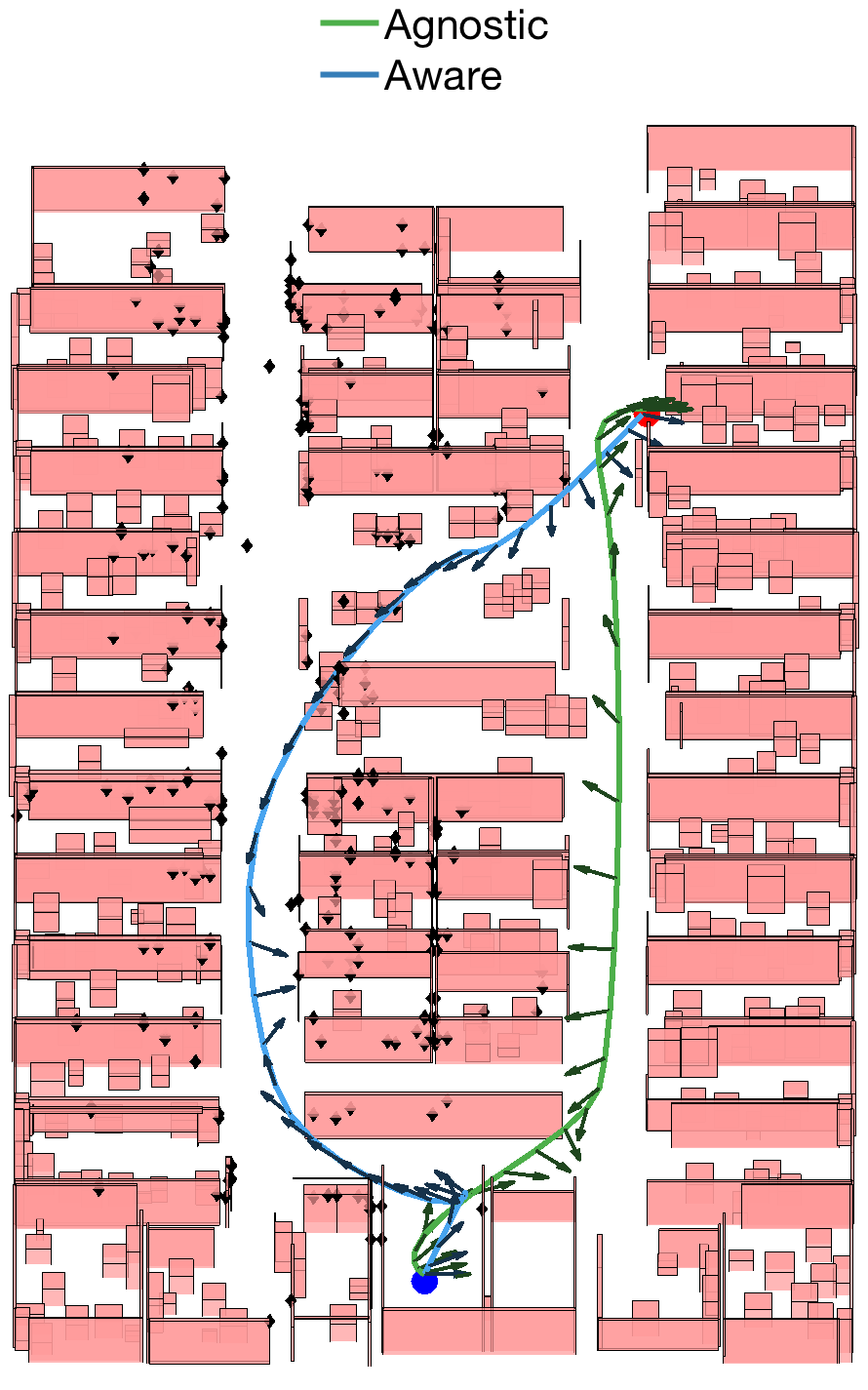}
        \caption{}
        \label{fig:gatesMap}
    \end{subfigure}
    \begin{subfigure}[b]{0.3\textwidth}
        \includegraphics[width=\textwidth]{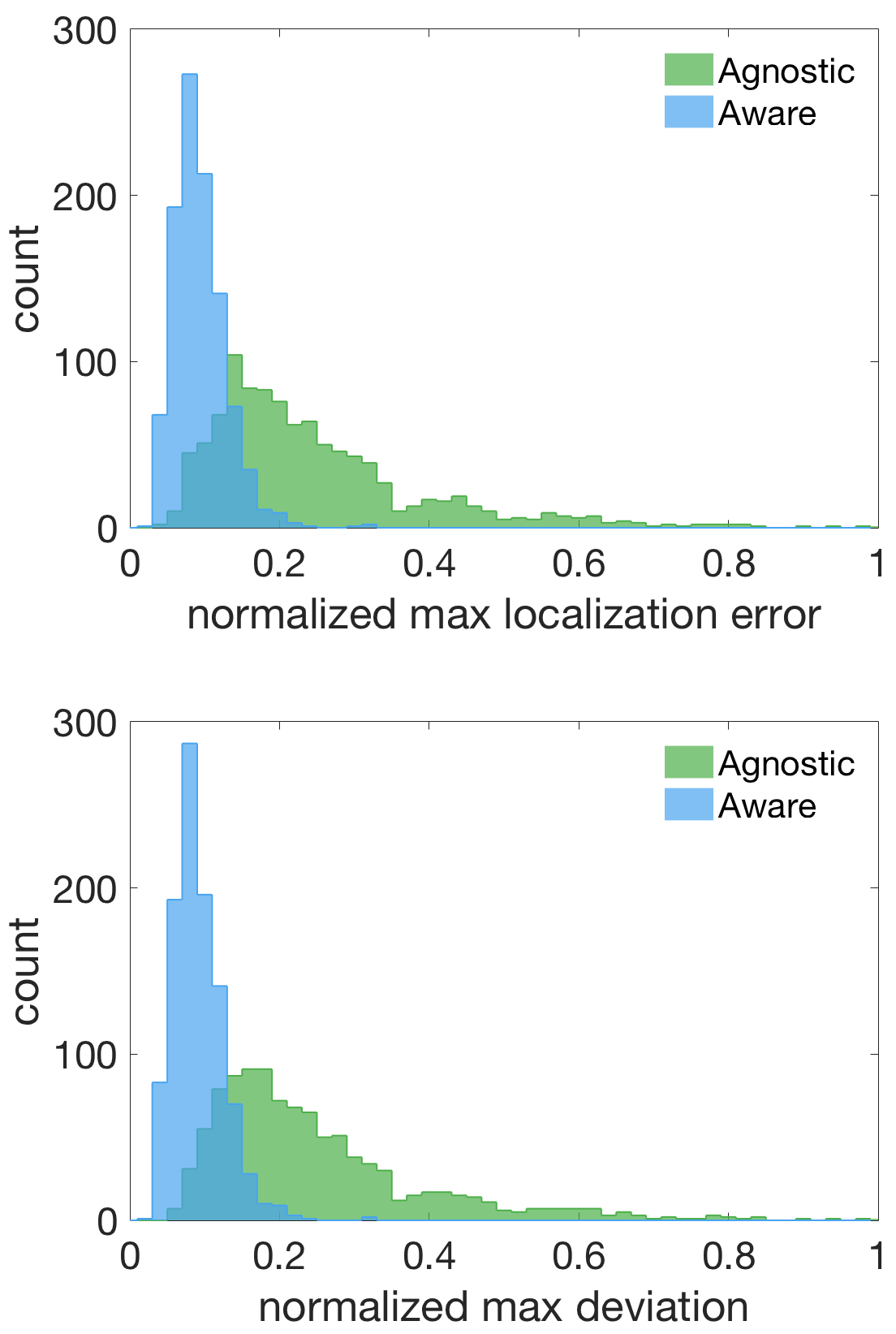}
        \caption{}
        \label{fig:gatesHist}
    \end{subfigure}
\caption{
(\ref{fig:gatesReal}) The point cloud representing the environment \cite{ArmeniSenerEtAl2016}.
(\ref{fig:gatesMap}) The obstacle representation of the environment as well as perception-aware and perception-agnostic trajectories.
Mapped features are shown in black, but only visible to a plan if within the field of view and unobstructed.
(\ref{fig:gatesHist}) The histograms of the distribution of localization error and deviation for each trajectory, generated via MC simulations. The results show both lower mean deviations and significantly lower variances for the perception-aware algorithm. The distributions vary especially in the right-tails.
}
\label{fig:gates}
\end{figure}

The second simulation environment, shown in Fig.~\ref{fig:gates}, represents a more complex, realistic environment generated from \cite{ArmeniSenerEtAl2016}, where individual objects are bounded by axis-aligned boxes.
The problem's initial state and goal region require the planner to compute trajectories for the full size of the building.
The features were generated in this environment to favor the high-cost hallway, representing an incomplete mapping in the low-cost hallway.
The resulting trajectories show the perception-agnostic formulation plans a more direct trajectory, while the perception-aware formulation plans a more robust trajectory at a higher cost.
Corroborating the results in the other obstacle set, the mean and variance of the perception-aware trajectory deviations are lower, but the largest difference occurs in the right tails of the distributions (Table~\ref{table:costDev}).

Lastly, Table~\ref{table:num} details the computation time for each phase and the cost over a range of selected sample counts.
On average over all runs, the computation time is allocated as 1\% graph building, 65\% perception heuristic computation, 7\% exploration, and 27\% MC.
This further motivates the use of an online implementation discussed at the end of Section~\ref{sec:alg} to offset the computation cost of the perception heuristic computation.
For other cases when the environment is known a priori (as is assumed in this work), the computation time reduces to the multiobjective search and potential MC verification.
The MC trials too can be computationally intensive due to the repeated computations of field of view, though we note our current implementation is completely unoptimized and considering a trajectory assumed fully recomputing features at 50~Hz.

\begin{table}[h!tbp]
\begin{center}
\begin{tabular}{|r| C{1.2cm} C{1.2cm} C{1.2cm} | C{1.2cm} C{1.2cm} |}
	\cline{2-6}
	\multicolumn{1}{c|}{} & \multicolumn{3}{ c |}{Fig.~\ref{fig:input}} & \multicolumn{2}{c |}{Fig.~\ref{fig:gates}} \\
	\hline
  Perception-Awareness & High & Low & Agnostic & High & Agnostic \\
    \hline
  Max Localization Error in Top 1\% of Plans & 0.30 & 0.37 & 0.83 & 0.21 & 0.79 \\
  Max Deviation in Top 1\% of Plans & 0.21 & 0.59 & 0.84 & 0.21 & 0.79 \\
  Cost & 288 & 211 & 201 & 263 & 193 \\
	\hline
  \end{tabular}
\end{center}
\caption{Simulation results over varying levels of perception-awareness. Note the max error and deviation are normalized by the max, max error and deviation observed over all 1000 trials. Costs are provided for comparison between levels of perception-awareness.}
\label{table:costDev}
\end{table}

\vspace{-0.2cm}
\begin{table}[h!tbp]
\begin{center}
\begin{tabular}{|r| C{1.2cm} C{1.2cm} | C{1.2cm} C{1.2cm} |}
	\cline{2-5}
	\multicolumn{1}{c |}{} & \multicolumn{2}{c |}{Fig.~\ref{fig:input}} & \multicolumn{2}{c |}{Fig.~\ref{fig:gates}} \\
	\hline
  	Sample Count & 2k & 4k & 2k & 4k \\
    \hline
  	Cost & 296 & 288 & 302 & 263\\
    Total Time (ms) & 49 & 147 & 812 & 1200 \\
    \scriptsize{\emph{Graph Building (Alg.~\ref{alg:graphBuild})}} & \scriptsize{\emph{1}} & \scriptsize{\emph{2}} & \scriptsize{\emph{6}} & \scriptsize{\emph{20}} \\
    \scriptsize{\emph{Perception Heuristic}} & \scriptsize{\emph{34}} & \scriptsize{\emph{116}}  & \scriptsize{\emph{276}} & \scriptsize{\emph{919}} \\
    \scriptsize{\bfseries{Exploration (Alg.~\ref{alg:explore})}} & \scriptsize{\bfseries{6}} & \scriptsize{\bfseries{19}} & \scriptsize{\bfseries{2}} & \scriptsize{\bfseries{14}} \\
    \scriptsize{\emph{Monte Carlo}} & \scriptsize{\emph{8}} & \scriptsize{\emph{10}} & \scriptsize{\emph{528}} & \scriptsize{\emph{247}} \\
	\hline
  \end{tabular}
\end{center}
\caption{Perception-aware plan computation times and costs (reported to demonstrate the trend of lower-cost trajectories with more samples).
In practice, since the environment can be mapped a priori, the full run time can be as little as just the exploration step.
Note the trajectory in Fig.~\ref{fig:gates} traverses the full building and represents an upper bound on what we expect the planner to compute.
}
\label{table:num}
\end{table}

\section{Physical Experiments}\label{sec:exp}

\subsection{Experimental Testbed}\label{sec:exptestbed}

The quadrotor used for the experiment was based on the DJI F330 frame (Fig.~\ref{fig:expQuad}). A Pixhawk autopilot running the PX4 software stack \cite{PXDT} was used for flight control. 
Additionally, the platform was augmented with a Linux computer, the Odroid XU4, responsible for communicating with the Pixhawk and bridging the autopilot with the rest of the network through the Robot Operating System (ROS). 
Using ROS, the platform exposes an interface for a ground-station to send position waypoints to the quadrotor. Trajectories are flown using this waypoint interface. 
The quadrotor was also equipped with a Google Tango smartphone. 
Tango performs a feature-based visual-inertial odometry along with a feature-based loop closure (termed ``area learning''); note the exact features and algorithmic methods are not known to the authors and thus the MC step was not performed. 
The localization computed by Tango was broadcasted over the network using ROS, transformed appropriately, and then used as a sensor measurement in the autopilot's Extended Kalman Filter.

The quadrotor was flown in a room measuring approximately 3m $\times$ 3m $\times$ 4m equipped with motion tracking cameras to provide ground-truth measurements, i.e., not used by the quadrotor for navigation. Instead, they are only used to evaluate and learn the performance of the Tango position estimate and train our heuristic. The room was learned with Tango's area learning feature with a large obstacle made up of a white table cloth in the middle of the room along with two additional sheets on the walls on one side of the room (Fig.~\ref{fig:expPic}). 
The presence of white uniform sheets reduces the number of visual features available to Tango in those regions. 
Moreover, the wind disturbances generated by the propellers tend to make the sheets move, resulting in non-stationary features. 
This makes it even more difficult for Tango to accurately track features that it can reliably use for pose estimation.

\subsection{Learned Heuristic}\label{sec:learnedheur}

\begin{figure}[b]
    \centering
    \begin{subfigure}[b]{.53\textwidth} 
        \includegraphics[width=\textwidth]{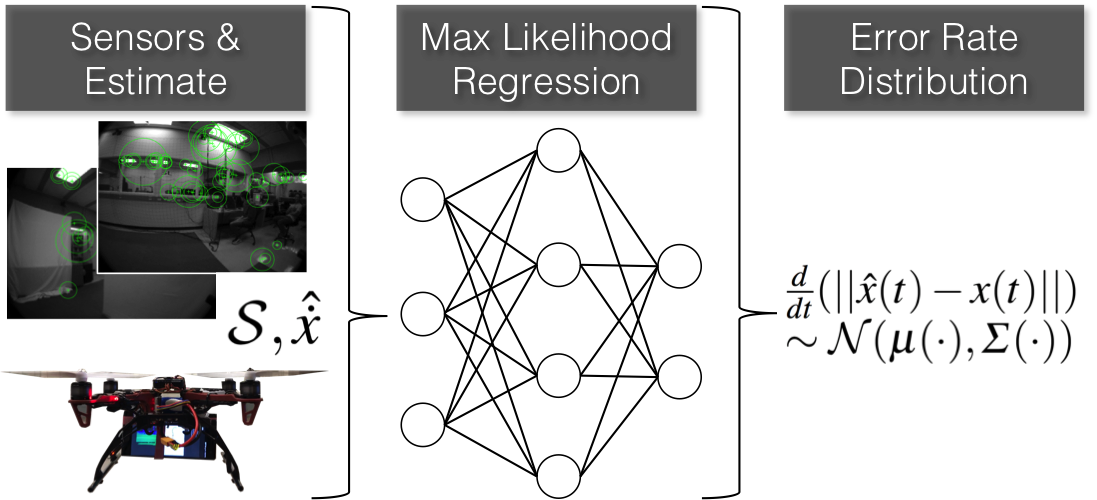}
        \caption{}
        \label{fig:learnedHeuristic}
    \end{subfigure}
    \begin{subfigure}[b]{.45\textwidth} 
        \includegraphics[width=\textwidth]{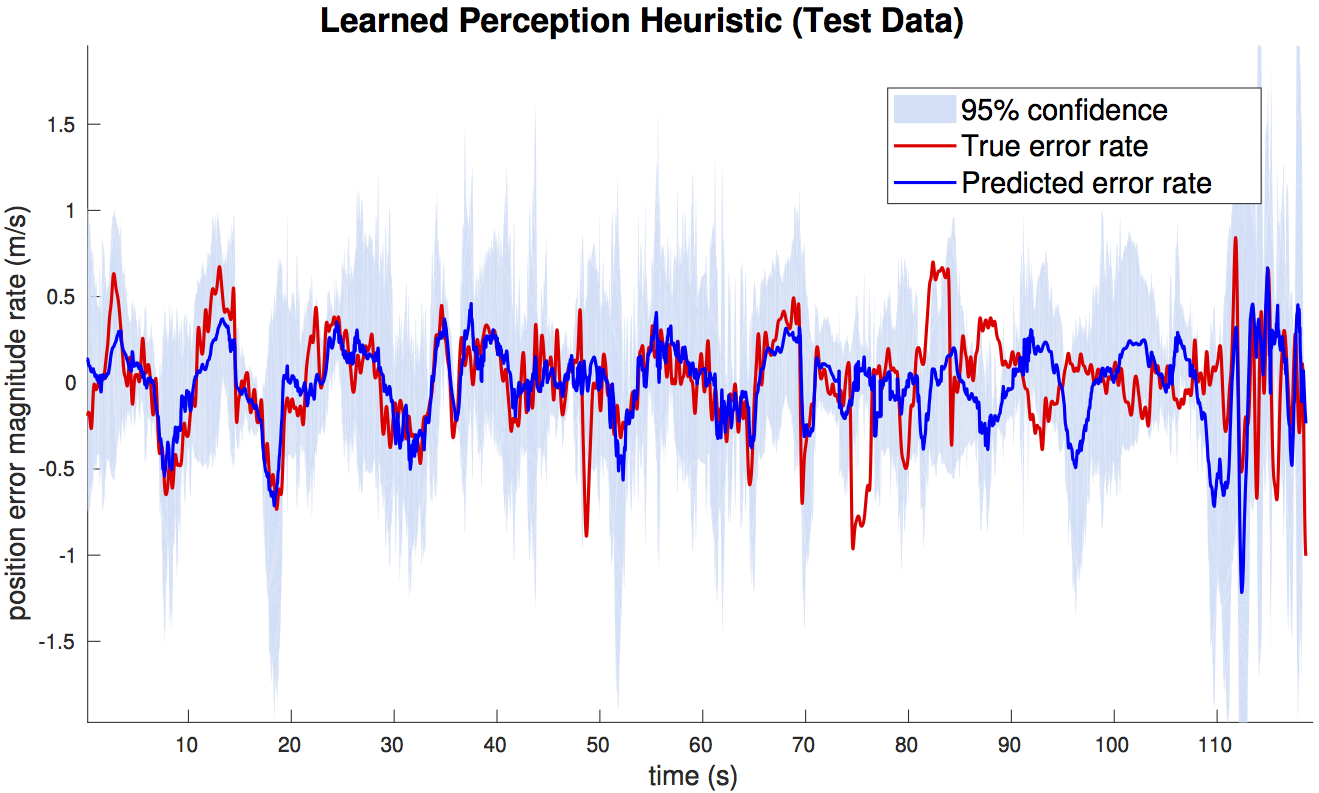}
        \caption{}
        \label{fig:heuristicTrain}
    \end{subfigure}
    \caption{
    (\ref{fig:learnedHeuristic}) The framework of our learned heuristic, the localization error rate distribution is predicted from the system's sensor output $\mathcal{S}$ and estimated rates $\hat{\dot{x}}$.
    (\ref{fig:heuristicTrain}) Rate of the position error magnitude predicted by the learned heuristic over test data. The distribution at each timestep is predicted independently. The data was generated by flying Google Tango inside of our motion capture system and evaluating the deviation between the two.}
    \label{fig:heuristicfig}
\end{figure}

We consider localization through Google Tango to be a black box, due to its software complexity and the uncertain nature of its loop-closure update, and propose a learning-based approach in order to derive the perception heuristic used in our experiments. This framework is outlined in Fig~\ref{fig:learnedHeuristic}. We acquired data on the perception quality of Google Tango by flying the quadrotor inside a motion capture system and computing the error between the two state estimates.
We make the modeling assumption that the position error rate produced by Google Tango (i.e., the rate of error magnitude between the onboard estimate and ground truth) is drawn from a time-varying Gaussian distribution, and fit the parameters of this distribution at each timestep/image frame using a maximum likelihood objective.

We trained a deep neural network for this distributional prediction, with the linear and angular velocities of the quadrotor and the number of SURF features visible by the camera as learning inputs. We did not include position as a training feature, because it would prevent the heuristic from generalizing outside of the motion capture room.
The neural network is a feedforward network with two hidden layers. The hidden layers contain eight nodes each and rectified linear unit activation functions. A final projection layer produces the mean and standard deviation of the instantaneous error rate distribution. Example test output of the prediction network is shown in Fig.~\ref{fig:heuristicTrain}; we note that the error rate model performs quite well within the confines of our motion capture arena. We leave to future work the validation of this learned heuristic in other environments with ground truth state measurement.

When mapping a new environment, the neural network is evaluated along the way using the number of SURF features visible as well as the current linear and angular velocities of the quadrotor, resulting in a mapping of the heuristic throughout the state space. This map is then interpolated using weighted nearest-neighbors (over the position, velocity, and yaw of the quadrotor) in order to produce the value of the heuristic for unvisited states (both at samples and along edges) used during the graph building phase of MPAP. A contour plot of this procedure applied to the experimental environment is shown in Fig.~\ref{fig:contour}, where the region surrounded by white sheets predicts poor localization performance, as expected.

\subsection{Experimental Results}\label{sec:expresults}

\begin{figure}[H]
    \centering
    \begin{subfigure}[b]{.44\textwidth}
        \includegraphics[width=\textwidth]{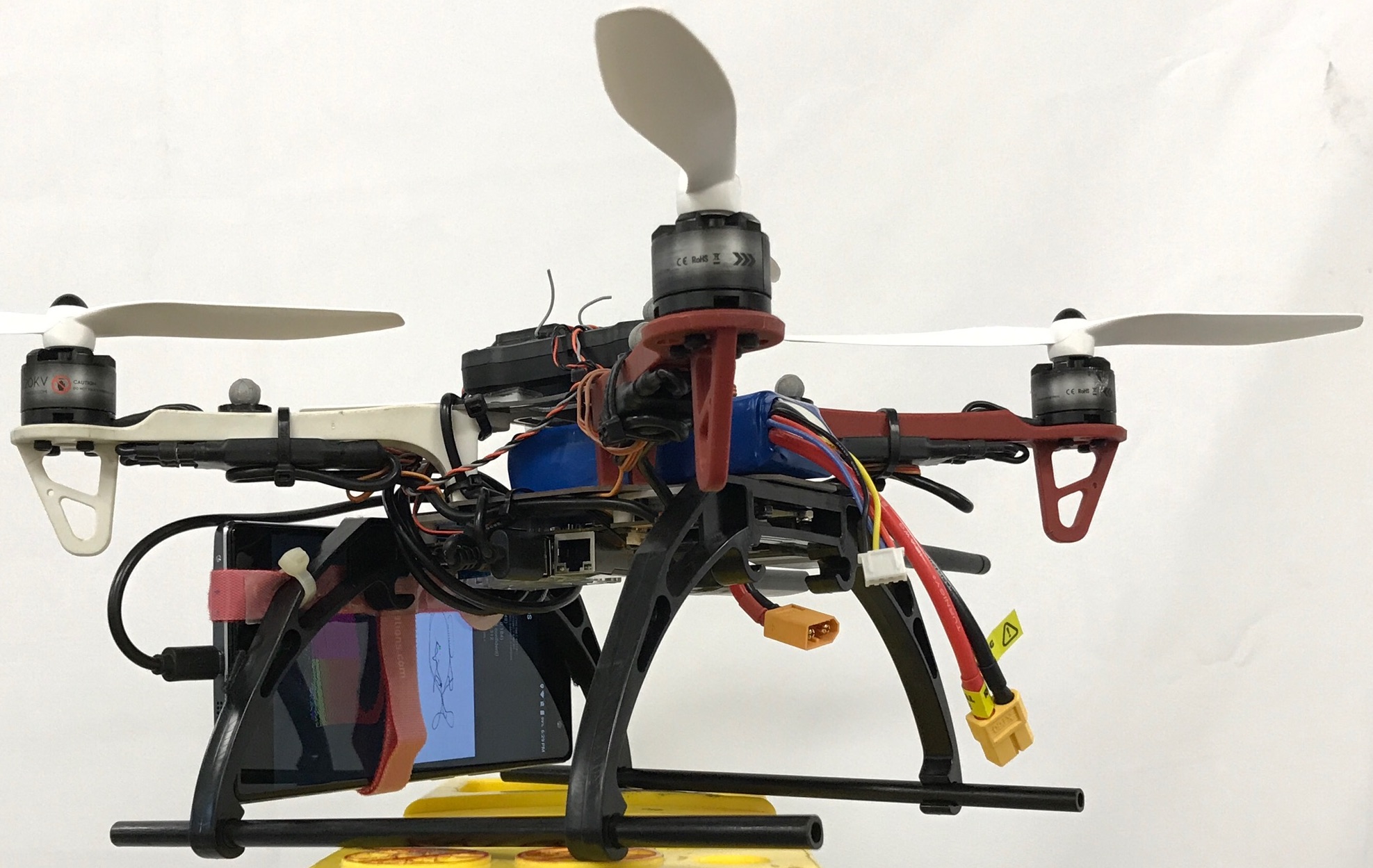}
        \caption{}
        \label{fig:expQuad}
    \end{subfigure}
    \begin{subfigure}[b]{.545\textwidth}
        \includegraphics[width=\textwidth]{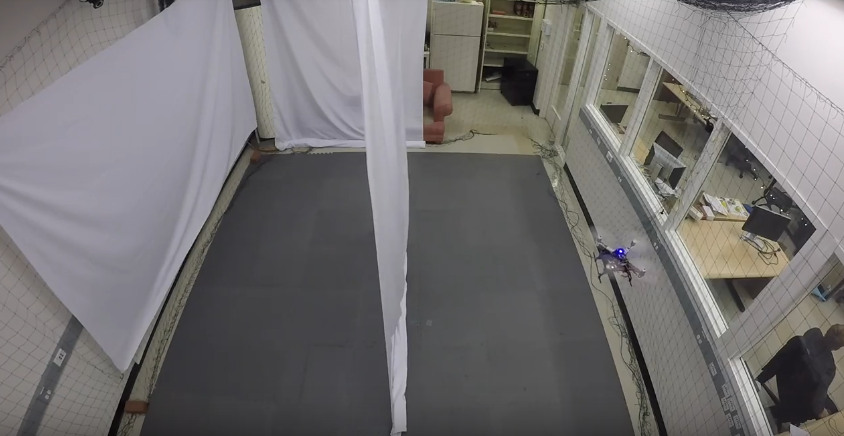}
        \caption{}
        \label{fig:expPic}
    \end{subfigure}
    \begin{subfigure}[b]{.435\textwidth}
        \includegraphics[width=\textwidth]{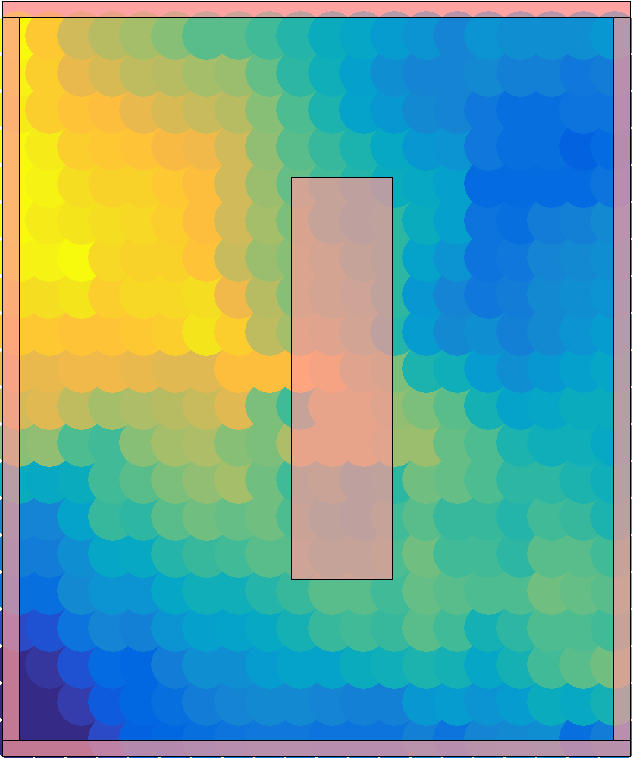}
        \caption{}
        \label{fig:contour}
    \end{subfigure}
    \begin{subfigure}[b]{.55\textwidth}
        \includegraphics[width=\textwidth]{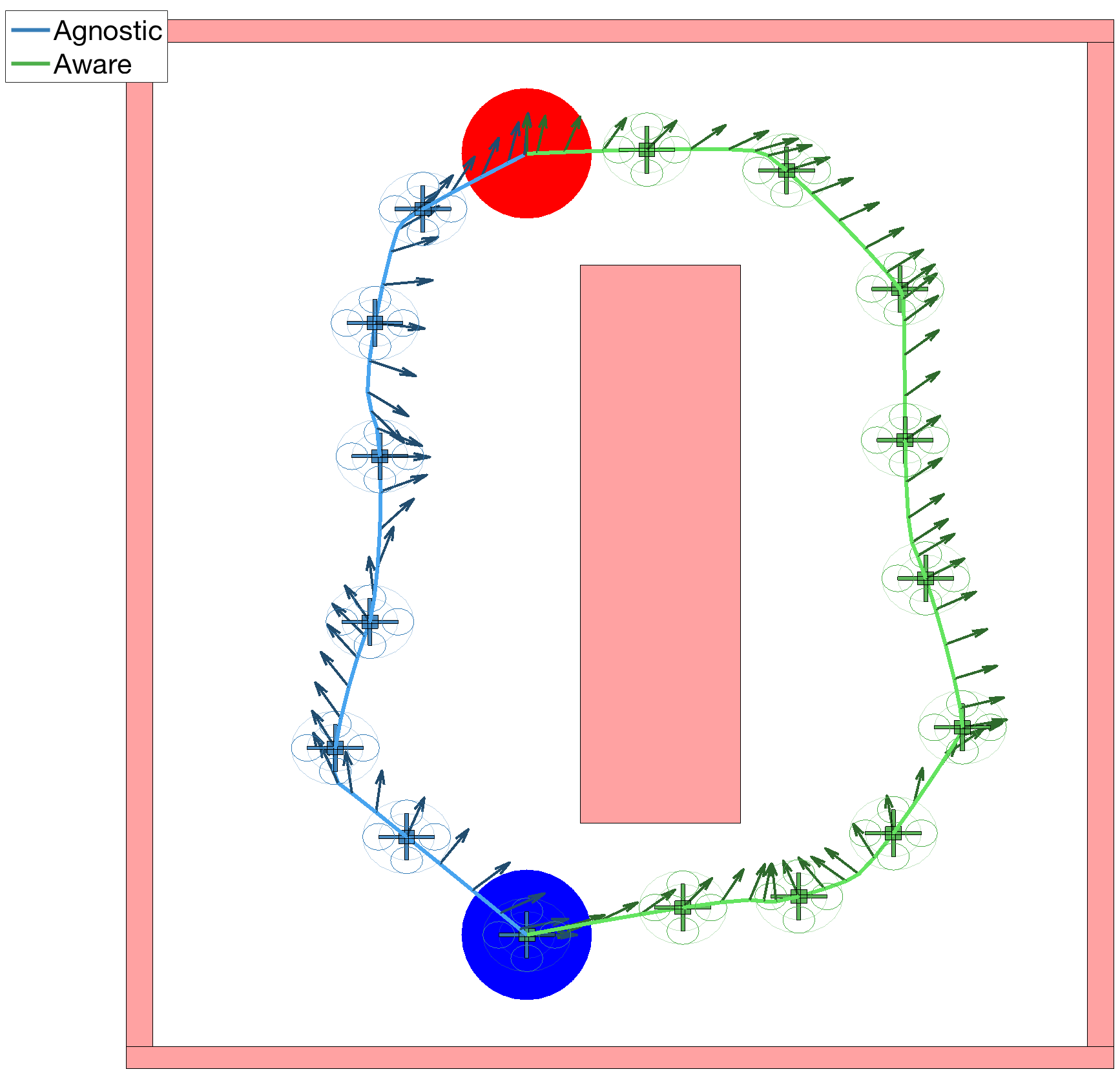}
        \caption{}
        \label{fig:expPlanned}
    \end{subfigure}
    \begin{subfigure}[b]{.47\textwidth}
        \includegraphics[width=\textwidth]{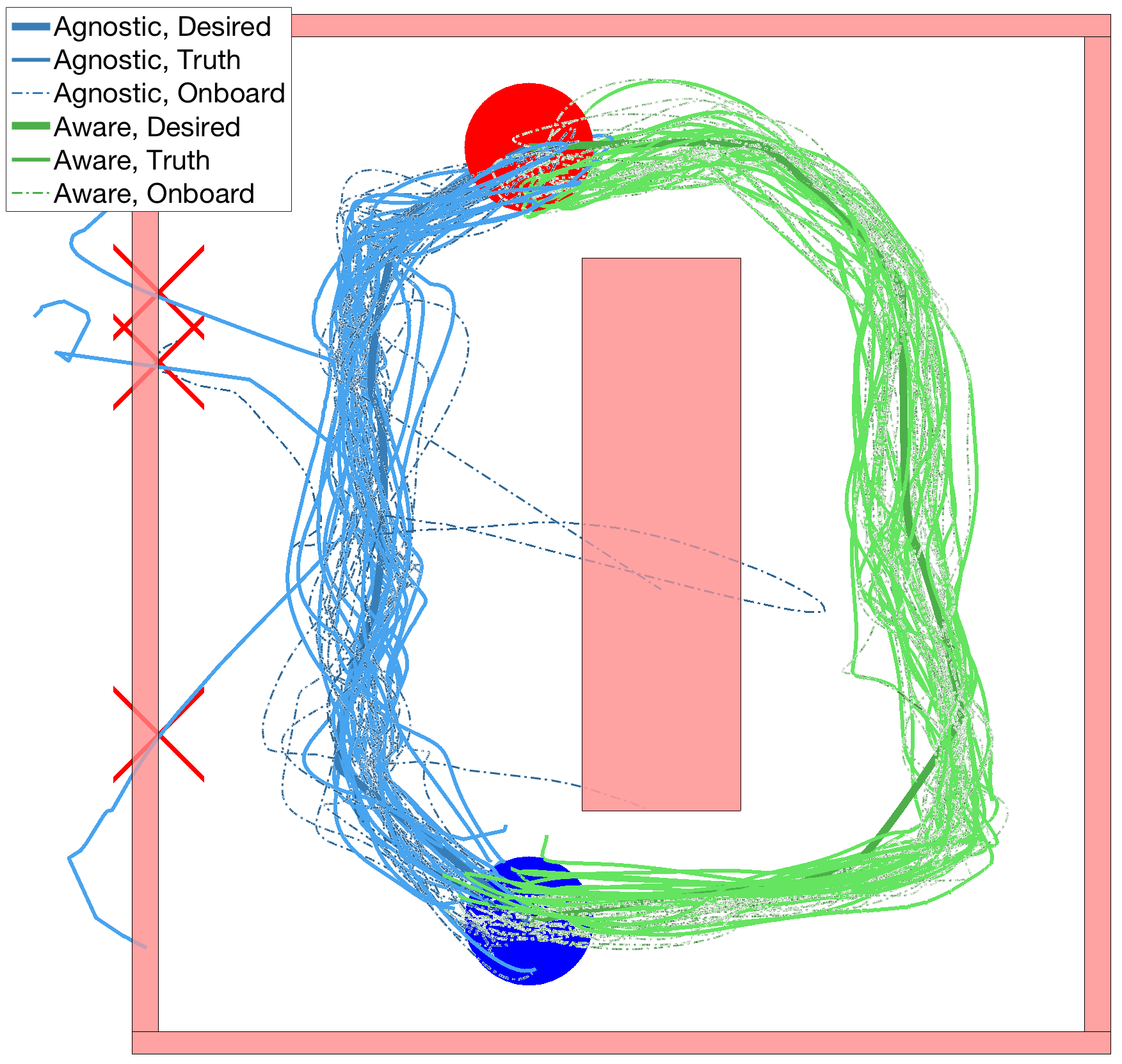}
        \caption{}
        \label{fig:expResults}
    \end{subfigure}
    \begin{subfigure}[b]{0.51\textwidth}
        \includegraphics[width=\textwidth]{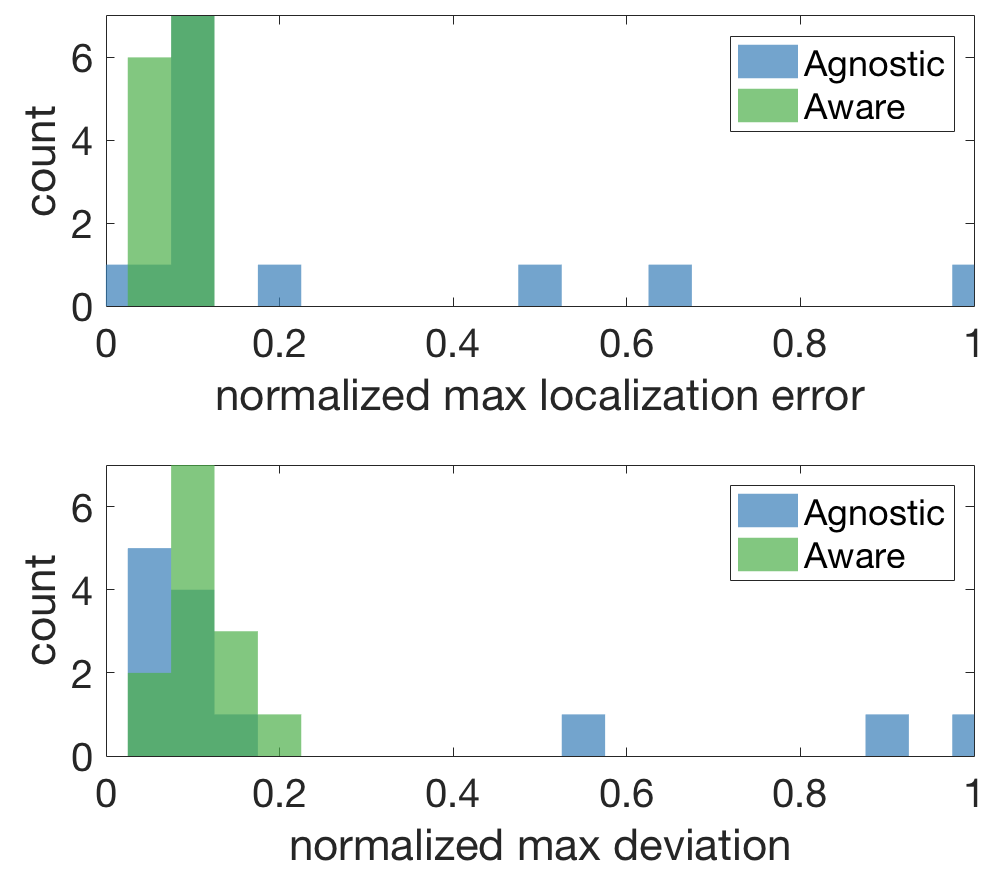}
        \caption{}
        \label{fig:expResultsHist}
    \end{subfigure}
\caption{
(\ref{fig:expQuad}) Quadrotor with mounted Google Tango used in experiments.
(\ref{fig:expPic}) Experimental environment setup, with white sheets on the left half of the room to reduce features.
(\ref{fig:contour}) Nearest neighbor contour plot from the learned heuristic (each position is averaged over a suite of yaw directions and zero velocity), where yellow represents poor localization and blue represents good. Note the region surrounded by white sheets is predicted to perform poorly.
(\ref{fig:expPlanned}) Planned trajectories, where known feature rich regions are represented with black diamonds.
(\ref{fig:expResults}) Desired, ground truth (Vicon), and onboard (Tango) estimate trajectories for the quadrotor over 26 flights, including 3 crashes for the perception-agnostic trajectory.
(\ref{fig:expResultsHist}) Histograms of maximum localization error and deviation in experiments.
}
\label{fig:exp}
\end{figure}

Two trajectories with the same initial and goal states (across the room from each other) were generated with the mapped environment (Fig.~\ref{fig:expPlanned}). 
One trajectory was computed using the full \mpap algorithm, with $\dr = 0.5$, while the other was computed as a baseline perception-agnostic trajectory.
Each trajectory was flown to the goal region and then back to the initial position. 

The motion plans were flown 13 times each with the resulting trajectories shown in Fig.~\ref{fig:expResults}. 
The perception-aware trajectory was able to fly all successfully, while the perception-agnostic trajectory crashed three times, all due to loss of localization events. 
These events are comparable to the long right tails observed in the numerical simulations in Section~\ref{sec:numexp}, Figs.~\ref{fig:inputHistEst} and \ref{fig:gatesHist}. 
As seen in Fig.~\ref{fig:expResultsHist}, outside of these localization losses, the two trajectories track the nominal approximately equally, despite the perception-aware being better localized. 
This is a result of the perception-aware plan incorporating more difficult turns for the controller to track. 
Videos of the two different plans being flown are available on the Stanford Autonomous Systems Laboratory Youtube channel.\footnote{https://www.youtube.com/watch?v=YKXf4iuj8-k}

\section{Conclusions}\label{sec:conc}

In this work we presented the Multiobjective Perception-Aware Planning algorithm to compute well-localized, robust motion plans. 
The algorithm leverages GPU massive parallelization to perform a multiobjective search (considering cost and a perception heuristic) over a graph representation of the state space, ultimately identifying a motion plan to be certified robust via Monte Carlo methods.
We demonstrated through numerical experiments that \mpap identifies motion plans in under a few seconds and that the objectives of perception and cost must be considered simultaneously to obtain truly robust plans.
Finally, we have shown through physical experiments that these perception-aware plans are indeed more robust and that the use of a learned perception heuristic, leveraging neural networks, can be effective in generating them.

This work leaves many avenues for further investigation.
First, we plan to implement the algorithm in a real-time onboard framework with simultaneous localization and mapping.
Second, we plan to extend our work on learned heuristics for planning, particularly towards heuristics that take the full sensor input directly.
Third, we plan to demonstrate this algorithm in more heterogeneous environments, e.g., outdoors, with varied lighting conditions.
Finally, while this work considered robustness through perception uncertainty, we plan to additionally consider environmental uncertainties and the ability of the controller to track trajectories while planning.

\newpage
\bibliographystyle{IEEEtran-short}
{
\renewcommand{\baselinestretch}{0.925}
\bibliography{../../../bib/main,../../../bib/ASL_papers}

\newcommand{\noopsort}[1]{} \newcommand{\printfirst}[2]{#1}
  \newcommand{\singleletter}[1]{#1} \newcommand{\switchargs}[2]{#2#1}
\begin{thebibliography}{10}
\providecommand{\url}[1]{#1}
\csname url@samestyle\endcsname
\providecommand{\newblock}{\relax}
\providecommand{\bibinfo}[2]{#2}
\providecommand{\BIBentrySTDinterwordspacing}{\spaceskip=0pt\relax}
\providecommand{\BIBentryALTinterwordstretchfactor}{4}
\providecommand{\BIBentryALTinterwordspacing}{\spaceskip=\fontdimen2\font plus
\BIBentryALTinterwordstretchfactor\fontdimen3\font minus
  \fontdimen4\font\relax}
\providecommand{\BIBforeignlanguage}[2]{{%
\expandafter\ifx\csname l@#1\endcsname\relax
\typeout{** WARNING: IEEEtran.bst: No hyphenation pattern has been}%
\typeout{** loaded for the language `#1'. Using the pattern for}%
\typeout{** the default language instead.}%
\else
\language=\csname l@#1\endcsname
\fi
#2}}
\providecommand{\BIBdecl}{\relax}
\BIBdecl

\bibitem{KaelblingLittmanEtAl1998}
L.~P. Kaelbling, M.~L. Littman, and A.~R. Cassandra, ``Planning and acting in
  partially observable stochastic domains,'' \emph{{Artificial Intelligence}},
  1998.

\bibitem{KurniawatiHsuEtAl2008}
H.~Kurniawati, D.~Hsu, and W.~S. Lee, ``{SARSOP}: Efficient point-based {POMDP}
  planning by approximating optimally reachable belief spaces,'' in
  \emph{{Robotics: Science and Systems}}, 2008.

\bibitem{PrenticeRoy2009}
S.~Prentice and N.~Roy, ``The belief roadmap: Efficient planning in linear
  {POMDPs} by factoring the covariance,'' \emph{{Int.\ Journal of Robotics
  Research}}, 2009.

\bibitem{BryRoy2011}
A.~Bry and N.~Roy, ``{R}apidly-exploring {R}andom {B}elief {T}rees for motion
  planning under uncertainty,'' in \emph{{Proc.\ IEEE Conf.\ on Robotics and
  Automation}}, 2011.

\bibitem{BergPatilEtAl2012}
J.~van~den Berg, S.~Patil, and R.~Alterovitz, ``Motion planning under
  uncertainty using iterative local optimization in belief space,''
  \emph{{Int.\ Journal of Robotics Research}}, 2012.

\bibitem{PatilKahnEtAl2014}
S.~Patil, G.~Kahn, M.~Laskeym, J.~Schulman, K.~Goldberg, and P.~Abbeel,
  ``Scaling up gaussian belief space planning through covariance-free
  trajectory optimization and automatic differentiation,'' in \emph{{Workshop
  on Algorithmic Foundations of Robotics}}, 2014.

\bibitem{IndelmanCarloneEtAl2015}
V.~Indelman, L.~Carlone, and F.~Dellaert, ``Planning in the continuous domain:
  A generalized belief space approach for autonomous navigation in unknown
  environments,'' \emph{{Int.\ Journal of Robotics Research}}, 2015.

\bibitem{BergAbbeelEtAl2011}
J.~van~den Berg, P.~Abbeel, and K.~Goldberg, ``{LQG-MP}: Optimized path
  planning for robots with motion uncertainty and imperfect state
  information,'' \emph{{Int.\ Journal of Robotics Research}}, 2011.

\bibitem{Agha-mohammadiAgarwalEtAl2016}
A.~{Agha-mohammadi}, S.~Agarwal, S.~Chakravorty, and N.~M. Amato,
  ``Simultaneous localization and planning for physical mobile robots via
  enabling dynamic replanning in belief space,'' \emph{{IEEE Transactions on
  Robotics}}, 2016, submitted, Available at
  \url{https://arxiv.org/abs/1510.07380}.

\bibitem{PlattTedrakeEtAl2010}
R.~Platt, R.~Tedrake, L.~Kaelbling, and T.~{Lozano-Perez}, ``Belief space
  planning assuming maximum likelihood observations,'' in \emph{{Robotics:
  Science and Systems}}, 2010.

\bibitem{AloimonosWeissEtAl1988}
J.~Aloimonos, I.~Weiss, and A.~Brandyopadhyay, ``Active vision,'' \emph{{Int.\
  Journal of Computer Vision}}, 1988.

\bibitem{ThrunBurgardEtAl2005}
S.~Thrun, W.~Burgard, and D.~Fox, \emph{Probabilistic Robotics}.\hskip 1em plus
  0.5em minus 0.4em\relax {MIT Press}, 2005.

\bibitem{CadenaCarloneEtAl2016}
C.~Cadena, L.~Carlone, H.~Carrillo, Y.~Latif, D.~Scaramuzza, J.~Neira, I.~Reid,
  and J.~Leonard, ``Past, present, and future of simultaneous localization and
  mapping: Towards the robust-perception age,'' \emph{{IEEE Transactions on
  Robotics}}, 2016.

\bibitem{SadatChutskoffEtAl2014}
S.~A. Sadat, K.~Chutskoff, D.~Jungic, J.~Wawerla, and R.~Vaughan,
  ``Feature-rich path planning for robust navigation of {MAVs} with
  {M}ono-{SLAM},'' in \emph{{Proc.\ IEEE Conf.\ on Robotics and Automation}},
  2014.

\bibitem{CostanteForsterEtAl2017}
G.~Costante, C.~Forster, J.~Delmerico, P.~Valigi, and D.~Scaramuzza,
  ``Perception-aware path planning,'' \emph{{IEEE Transactions on Robotics}},
  2017, submitted, Available at \url{https://arxiv.org/abs/1605.04151}.

\bibitem{CarloneLyons2014}
L.~Carlone and D.~Lyons, ``Uncertainty-constrained robot exploration: A
  mixed-integer linear programming approach,'' in \emph{{Proc.\ IEEE Conf.\ on
  Robotics and Automation}}, 2014.

\bibitem{IchterSchmerlingEtAl2017}
B.~Ichter, E.~Schmerling, A.~Agha-mohammadi, and M.~Pavone, ``Real-time
  stochastic kinodynamic motion planning via multiobjective search on {GPUs},''
  in \emph{{Proc.\ IEEE Conf.\ on Robotics and Automation}}, 2017.

\bibitem{KavrakiSvestkaEtAl1996}
L.~E. Kavraki, P.~{\v{S}}vestka, J.-C. Latombe, and M.~H. Overmars,
  ``Probabilistic roadmaps for path planning in high-dimensional spaces,''
  \emph{{IEEE Transactions on Robotics and Automation}}, 1996.

\bibitem{AmatoDale1999}
N.~M. Amato and L.~K. Dale, ``Probabilistic roadmap methods are embarrassingly
  parallel,'' in \emph{{Proc.\ IEEE Conf.\ on Robotics and Automation}}, 1999.

\bibitem{JansonSchmerlingEtAl2015b}
L.~Janson, E.~Schmerling, and M.~Pavone, ``{Monte} {Carlo} motion planning for
  robot trajectory optimization under uncertainty,'' in \emph{{Int.\ Symp.\ on
  Robotics Research}}, 2015.

\bibitem{DavisonMurray2002}
A.~J. Davison and D.~W. Murray, ``Simultaneous localization and map-building
  using active vision,'' \emph{{IEEE Transactions on Pattern Analysis \&
  Machine Intelligence}}, 2002.

\bibitem{ScaramuzzaFraundorfer2011}
D.~Scaramuzza and F.~Fraundorfer, ``Visual odometry part {I}: The first 30
  years and fundamentals,'' \emph{{IEEE Robotics and Automation Magazine}},
  2011.

\bibitem{LaValle2006}
S.~M. LaValle, \emph{Planning Algorithms}.\hskip 1em plus 0.5em minus
  0.4em\relax {Cambridge University Press}, 2006.

\bibitem{ArmeniSenerEtAl2016}
I.~Armeni, O.~Sener, A.~R. Zamir, H.~Jiang, I.~Brilakis, M.~Fischer, and
  S.~Savarese, ``{3D} semantic parsing of large-scale indoor spaces,'' in
  \emph{{IEEE Conf.\ on Computer Vision and Pattern Recognition}}, 2016.

\bibitem{PXDT}
{PX4 Development Team}, ``{PX4} autopilot,'' {Available} at
  \url{http://px4.io/}.

\end{thebibliography}
}

\end{document}